\documentclass[10pt,twocolumn,letterpaper]{article}

\usepackage{wacv}
\usepackage{times}
\usepackage{epsfig}
\usepackage{graphicx}
\usepackage{amsmath}
\usepackage{amssymb}
\usepackage{soul}
\usepackage{authblk}
\usepackage{booktabs}
\usepackage{bm}
\usepackage{subcaption}
\makeatother
\def\wacvPaperID{226} 

\wacvfinalcopy 
\ifwacvfinal
\usepackage[breaklinks=true,bookmarks=false]{hyperref}
\else
\usepackage[pagebackref=true,breaklinks=true,colorlinks,bookmarks=false]{hyperref}
\fi

\begin{document}

\title{Data-efficient Alignment of Multimodal Sequences by Aligning Gradient Updates and Internal Feature Distributions}

\author[1]{Jianan Wang}
\author[2,3]{Boyang Li}
\author[1]{Xiangyu Fan}
\author[1]{Jing Lin}
\author[1,4]{Yanwei Fu}
\affil[1]{School of Data Science, Fudan University, China}
\affil[2]{Nanyang Technological University, Singapore}
\affil[3]{Alibaba-NTU Singapore Joint Research Institute}
\affil[4]{MOE Frontiers Center for Brain Science, Fudan University, China
\authorcr{\tt\small boyangli@outlook.com}  \quad {\tt\small  jinglin0224@163.com} \quad {\tt\small \{jawang19, fanxy19, yanweifu\}@fudan.edu.cn}}

\maketitle
\thispagestyle{empty}

\begin{abstract}
The task of video and text sequence alignment is a prerequisite step toward joint understanding of movie videos and screenplays. However, supervised methods face the obstacle of limited realistic training data. With this paper, we attempt to enhance data efficiency of the end-to-end alignment network NeuMATCH \cite{NeuMatch2018}. Recent research \cite{wang2019makes} suggests that network components dealing with different modalities may overfit and generalize at different speeds, creating difficulties for training. We propose to employ (1) layer-wise adaptive rate scaling (LARS) to align the magnitudes of gradient updates in different layers and balance the pace of learning and (2) sequence-wise batch normalization (SBN) to align the internal feature distributions from different modalities. Finally, we leverage random projection to reduce the dimensionality of input features. On the YouTube Movie Summary dataset, the combined use of these technique closes the performance gap when the pretraining on the LSMDC dataset is omitted and achieves the state-of-the-art result. Extensive empirical comparisons and analysis reveal that these techniques improve optimization and regularize the network more effectively than two different setups of layer normalization.  
\end{abstract}

\section{Introduction}

\begin{figure*}[th]
\centering
\includegraphics[width=\textwidth]{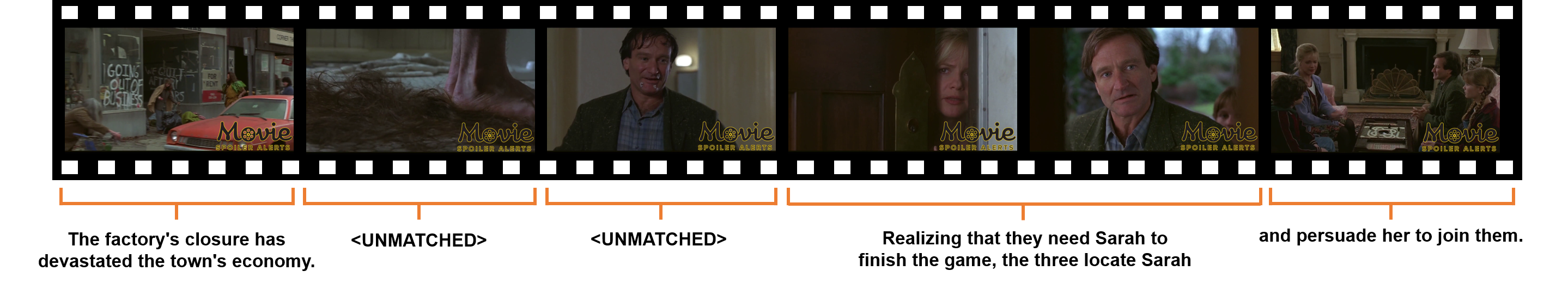}
\caption{An example of aligned video and text sequences from YMS~\cite{NeuMatch2018}. One text snippet can match multiple video clips and some video clips may be unmatched. }
\label{fig:alignment-task}
\end{figure*}

Today we have access to a massive amount of parallel video and textual data in the form of movie videos and screenplays. For instance, more than 1,200 movie scripts are available at the Internet Movie Script Database\footnote{\url{https://www.imsdb.com/}} under fair use. However, in their natural form, the data lack fine-grained correspondences at the level of sentences and video segments. Establishing such cross-modality correspondences will enable a variety of applications such as acquisition of multi-modal knowledge \cite{Tandon2015:Knowlywood,Wu_2019_CVPR,shi2019visually,wu2020analogical,YeGuangnan2015}, generation of one modality from the other \cite{rimle2020enriching,luo2020univl,wang2018reconstruction,Zhou_2018_CVPR,Wang2018:HRL-video-caption,Huijuan2019,li2017video,balaji2019conditional,LiuYue2019}, movie video retrieval \cite{Pavel2015}, and visualization of movie content  \cite{Kim2018}.
For this purpose, we investigate the alignment video and textual sequences (see an example of aligned sequences in Figure \ref{fig:alignment-task}).

A major obstacle of this task is the limited amount of training data due to the high cost of manually annotating sentence-level alignment. The Large Scale Movie Description Challenge (LSMDC) dataset \cite{LSMDC2017} 
is massive (158 hours of video and 124k texts), but it lacks one-to-many matching, unmatched elements, and imprecise language that are common in real-world data. The YouTube Movie Summary (YMS) dataset \cite{NeuMatch2018} is more realistic but is more than 20 times smaller (6.7 hours of video and 5.9k texts). 

For its small size, fast pace, complex correspondence relations, and storytelling language, the YMS dataset is a difficult challenge for training from scratch. In \cite{NeuMatch2018}, the end-to-end sequence alignment network NeuMATCH is pretrained on LSMDC and finetuned on YMS. In our experiments, directly training NeuMATCH on YMS unsurprisingly results in overfitting and a 2.4\% performance drop in text accuracy despite early stopping. 

\renewcommand{\arraystretch}{1.1}

With this paper, we aim to enhance the data efficiency of video-text sequence alignment and mitigate overfitting when training on small data. Recent research \cite{wang2019makes} suggests that in multimodal neural networks, sub-networks dealing with different modalities overfit and generalize at different speeds. This is consistent with our observation that applying regularization uniformly to a network that already overfits can sometimes exacerbate overfitting. To overcome this issue, we propose to (1) align the magnitudes of gradient-based parameter updates so that different modalities learn at similar rates and (2) align the internal feature distributions from different modalities.

To accomplish the first goal, we adopt layer-wise adaptive rate scaling (LARS), which is traditionally reserved for training with large batches \cite{LARS2017,you2019large}. In our experiments, the optimization of the four network stacks are more balanced under LARS than Adam (see Section \ref{sec:lars-analysis}), suggesting LARS helps in properly pacing learning in different network components . This work may be considered as redirecting the existing technique of LARS to a novel objective. To accomplish the second goal, we adopt sequence-wise batch normalization (SBN) \cite{SBN2015}, which normalizes the internal feature distributions from different modalities. We show that SBN is superior to a popular configuration of layer normalization \cite{ba2016:LayerNorm}, which normalizes the internal representations of LSTMs instead of outputs. 

Large pretrained video-text models (e.g., \cite{sun2019videobert,Zhu2020:ActBERT,miech2019endtoend,luo2020univl}) can serve as powerful feature extractors for single video clips and single sentences. However, the NeuMATCH network contains video/text \emph{sequence} encoders (a.k.a. the Video Stack and the Text Stack) that capture contextual information from long sequences. On average, they encode 180 clips and 66 text snippets for one video in YMS. It is not immediately clear how to apply existing methods to pretrain the sequence encoders. 
To handle potential overfitting in the sequence encoders, we propose to use random projection (RP)  \cite{Achlioptas2001:RP} to reduce the dimensionality of input features. 
After applying these techniques, we surpass the previous best performance with pretraining as well as a baseline with LXMERT \cite{tan2019lxmert} features, which is directly trained on YMS.

\noindent \textbf{Contributions.} First, we successfully improve the data efficiency of multimodal sequence alignment with three techniques, LARS, SBN, and RP. As a result, we achieve state-of-the-art results on the YMS dataset without pretraining on LSMDC. Second, by providing detailed empirical analysis of these techniques (Sections \ref{sec:lars-analysis}), \ref{sec:sbn-analysis}), and \ref{sec:rp-analysis})), we hint at their broader potential for improving data efficiency in multimodal tasks. To our knowledge, the combined use of the three techniques in multimodal tasks is novel.

\section{Related Work}

\noindent \textbf{Alignment of Video and Text Sequences.} 
The alignment of video and text sequences has long been a problem of research interest~\cite{cour2008movie, everingham2006hello}. A popular theme of solutions is metric learning followed by dynamic time warping, which can be understood as inference on a linear conditional random field (CRF) \cite{sankar2009subtitle, tapaswi2014story, tapaswi2015book2movie}. \cite{bojanowski2015weakly} adopts a constrained quadratic integer programming formulation. 
\cite{zhu2015aligning} is an early approach utilizing convolutional neural networks that consider a local context in learning similarity between video segments and book chapters. After that, a CRF approach is used for decoding. By decomposing the alignment into a sequence of action classification, NeuMATCH \cite{NeuMatch2018} provides the first end-to-end differentiable solution. However, due to the complexity and heterogeneity of the network, training can become a challenge. 

To deal with data sparsity, researchers turn to weakly supervised or unsupervised learning.
\cite{Huang2019:d3tw} utilizes the sequential constraints for weakly supervised action localization and proposes a differentiable dynamic time warping objective function. \cite{Yu_sentence_2017} discovers objects from textual descriptions of their locations and movements. \cite{naim2015:unsup-align} is an unsupervised approach for aligning textual instructions and demonstration videos based on object names and images. 

\vspace{0.1in}
\noindent \textbf{Text Grounding in Video.} Temporal grounding of text, also called moment retrieval or temporal activity localization, seeks a temporal interval of the video that corresponds to a textual query \cite{Zhang_2019_CVPR}. This is in contrast to spatio-temporal grounding, where texts are grounded in spatio-temporal tubes \cite{chen2019weaklysupervised}, and the spatial grounding (\emph{e.g.}, \cite{Dogan2019:SeqGround,wang2019phrase}), where texts are grounded in spatial locations of images. 
Here we briefly review fully supervised temporal grounding. TALL \cite{gao2017tall} compares the sentence to sliding-window clip proposals and tweaks their temporal locations. \cite{hendricks-etal-2018-localizing} learns to rank the correct proposal higher than incorrect ones. \cite{xu2018multilevel} uses a proposal network to predict the proposal's location directly. \cite{chen-etal-2018-temporally} formulates the task as sequence labeling, where the classification at each time step selects among proposals of predefined lengths ending at the current time. \cite{He2019:grounding-RL,Wu2020:tree-structured-RL} introduce reinforcement learning methods.

\vspace{0.1in}
\noindent \textbf{Temporal Activity Proposal.} 
The proposal network, first used in object detection \cite{Ren2015:Faster-RCNN}, is often used to identify potentially useful portions of video or image as the candidates for grounding. In the temporal grounding task, we conventionally place proposals of predefined lengths at all possible locations; proposals are classified as valid or invalid and their boundaries are adjusted \cite{Gao_2017_ICCV}. The frame-based approach in \cite{Lin2018bsn} finds likely starting and ending positions of proposals and collects candidate proposals in-between.  \cite{Liu_2019_CVPR} creates proposals of different lengths using a downsampling and upsampling strategy similar to U-Net \cite{U-Net2015}. \cite{Ji2019:proposals} introduces data augmentation techniques, time warping and time masking, in semi-supervised learning. 

The semantic content of a natural language sentence incurs considerably greater variations than action labels, which usually contain 1-3 words. Thus, it is challenging to  \emph{a priori} determine good candidates for sentence grounding. For this reason, we let one sentence to match with multiple predefined video clips rather than assuming it matches with exactly one proposal found by a proposal network. 

\vspace{0.1in}
\noindent \textbf{Pacing Multimodal Gradient Updates.} 
End-to-end optimization has been a driving force for the many successes of deep learning, yet it is only understood recently that interactions between gradients for different network components may interfere with learning \cite{letcher2019differentiable}. Wang et al. \cite{wang2019makes} employ a convex combination of unimodal gradients so that different modalities may learn at similar rates. However, the computation of unimodal gradients is not easily applicable in the sequence alignment task, which always requires input from two modalities. As an alternative, we propose to maintain the same ratio between the the gradient magnitudes and the layer weights, so that the paces of learning across network components may be aligned.

\begin{figure*}[th]
\centering
\includegraphics[scale=0.63]{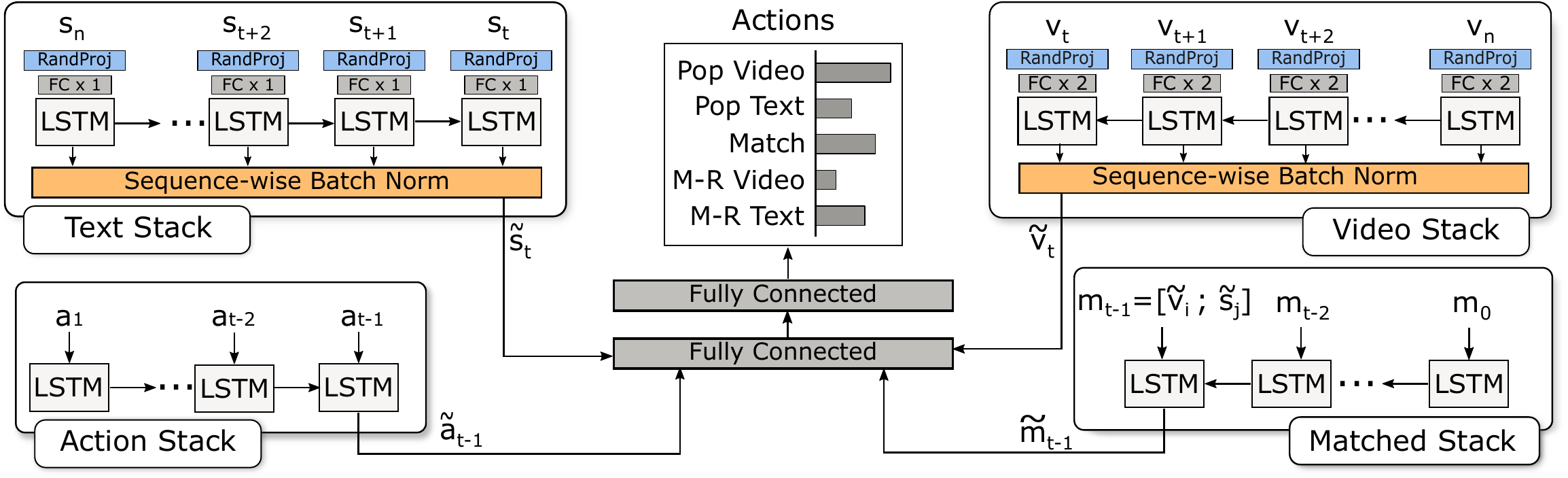}
\caption{The Network Architecture. New components proposed in this paper are shown in blue and orange backgrounds.}
\label{fig:architecture}
\end{figure*}

\section{Approach}

\subsection{NeuMATCH}
NeuMATCH \cite{NeuMatch2018} is an end-to-end differentiable neural network for multi-sequence alignment. The basic idea is to represent the current state of partially matched sequences as a vector and classify it into an action. Incrementally, the actions determine if the topmost elements in the two sequences should be matched or if they should be removed from consideration. After one decision, the second element may move to the top position and get processed. This process repeats until one of the sequences is exhausted. 

\vspace{0.1in}
\noindent  \textbf{Network Architecture.} Shown in Figure \ref{fig:architecture}, the NeuMATCH architecture employs four LSTM networks that respectively represent the video sequence, the text sequence, previously executed actions, and previously matched elements. The network only operates on the topmost elements on the four stacks. 

The LSTM network \cite{LSTM1999} is commonly applied to sequential data and can be characterized as follows. Let the sequence of input features be $\langle \bm x_1, \bm x_2, \ldots, \bm x_T \rangle$, or $\langle \bm x_t \rangle_{t=1}^T$ for short. LSTM outputs a sequence of hidden states $\langle \bm h_t \rangle_{t=1}^T$ and internal cell states $\langle \bm c_t \rangle_{t=1}^T$. For any $t \in \{1 \ldots T\}$, $\bm x_t, \bm h_t, \bm c_t \in \mathcal{R}^D$. In temporally forward LSTM, $\bm h_t$ is updated as follows.
\begin{equation}
\begin{bmatrix}
\bm i_t^\top, \bm f_t^\top, \tilde{\bm c}_t^\top, \bm o_t^\top 
\end{bmatrix}^\top = W_x \bm x_t + W_h \bm h_{t-1} + \bm b 
\end{equation}
\begin{equation}
\bm c_t = \sigma(\bm f_t) \otimes \bm c_{t-1} + \sigma(\bm i_t) \otimes \tanh(\tilde{\bm c}_t) 
\end{equation}
\begin{equation}
\bm h_t = \sigma(\bm o_t) \otimes  \tanh(\bm c_t)
\label{eq:lstm-h}
\end{equation}
where $W_{x}, W_{h}  \in \mathcal{R}^{4D\times D}$ and $\bm b \in \mathcal{R}^{4D}$ are trainable parameters. $\otimes$ denotes component-wise multiplication. Note that LSTM may run temporally backward, so that $\bm h_t$ becomes a function of $\bm x_{t}$, $\bm h_{t+1}$, and $\bm c_{t+1}$, as in the Video and Text Stacks. 

In the Video Stack and the Text Stack, the input sequence of video features is denoted as $\langle \bm v_i \rangle_{i=1}^N $ and the input text features are $\langle \bm s_j \rangle_{j=1}^M$. These features go through a random projection layer for dimensionality reduction (see Section \ref{sec:rand_proj}) and fully connected (FC) layers before the LSTM. At any time $t$, the topmost element in the two stacks are denoted as $\bm v_t$ and $\bm s_t$ respectively. 
To model contextual information from the rest of the sequence, the two LSTMs feed temporally backward. The LSTM hidden states go through sequence-wise batch normalization (see Section \ref{sec:sbn}), whose normalized output are $\langle \tilde{\bm v}_i \rangle_{i=1}^N$ and $\langle \tilde{\bm s}_j \rangle_{j=1}^M$. 

The Action Stack, feeding temporally forward, provides contextual information from previous actions executed by the network. Its input sequence contains one-hot vectors $\langle \bm a_k \rangle_{k=1}^{t-1}$ and its output is the last hidden state $\tilde{\bm a}_{t-1}$. Similarly, the Matched Stack provides context from previously matched video and text elements. Its input sequence $\langle \bm m_k \rangle_{k=1}^{t-1}$ contains the concatenation of video and text features. For example, if at time $k$, the $i^{\text{th}}$ video and the $j^{\text{th}}$ text are matched, we let $\bm{m}_k = [\tilde{\bm v}_i;\tilde{\bm s}_j]$. In case of one-to-many matching, we take the average of multiple elements in the same modality and the same slot. The Matched Stack outputs the last hidden state, $\tilde{\bm m}_{t-1}$. 

The network extracts an overall state vector $[\tilde{\bm v}_t; \tilde{\bm s}_t; \tilde{\bm a}_{t-1}; \tilde{\bm m}_{t-1}]$ from all four stacks. The state goes through two FC layers before a softmax function, where it is classified into one of the available actions as defined in the next paragraph. The network is trained using cross-entropy loss. 

\vspace{0.1in}
\noindent \textbf{Action Definitions.} We define five actions that can be used in the alignment of two sequences in a one-to-one or one-to-many setting with unmatched elements. The actions are Pop-Text, Pop-Video, Match, Match-Retain-Text, and Match-Retain-Video. Depending on the nature of the data, we may employ a subset of all possible actions.
The Pop-Text action removes the topmost element from the Video Stack so that the second element can move up and get aligned. Similarly, the Pop-Video action removes the topmost element from the Text Stack. The Match action matches the two topmost elements, removes them from the Video and Text Stacks, and inserts them at the topmost position of the Matched Stack. These three actions are sufficient for one-to-one matchings. For the situation where one text element can match with more than one video element, the action Match-Retain-Text action removes the video element but leave the text element at the top of the Text Stack, so it may be matched with another video element. The matched elements are still inserted into the Matched Stack. The Match-Retain-Video action functions analogously. 

\subsection{Layer-wise Adaptive Rate Scaling (LARS)}
Layer-wise Adaptive Rate Scaling (LARS) \cite{LARS2017} scales the gradient updates for different layers so that the ratios between the update and the parameter magnitudes remain the same across layers. More formally, let $\bm w^l$ denote the parameters for the $l^{\text{th}}$ layer and $\nabla_{w^l}\mathcal
{L}$ denote the gradient of the loss function  with respect to $\bm w^l$. The update $\Delta \bm w^l$ is calculated as
\begin{equation}
    \Delta \bm w^l = \eta \|\bm w^l\|_2  \frac{\nabla_{\bm w^l}\mathcal {L}}{\|\nabla_{\bm w^l}\mathcal
{L}\|_2} 
\label{eq:lars}
\end{equation}
where $\eta$ is a global learning rate and $\|\cdot\|_2$ denotes the $\ell_2$ norm. It is easy to see that, under LARS, $\Delta \bm w^l$ takes the direction of $\nabla_{w^l}\mathcal
{L}$ and has the magnitude of $\eta \|\bm w^l\|_2$. Thus, LARS ensures that the update is proportional to the parameters weights, so that layers with poorly scaled gradients can be optimized. In this paper, we apply LARS to gradient updates computed by Adam and replace the gradient term $\nabla_{\bm w^l}\mathcal{L}$ in Eq. \ref{eq:lars} with the Adam update. LARS applies layer-wise scaling of the gradient whereas Adam applies component-wise scaling, obtaining complementary effects. 

The effects of LARS have been mostly reported in large-batch training \cite{LARS2017,you2019large}. In this paper, we utilize LARS to align the updates between network components dealing with different modalities. 

\subsection{Sequence-wise Batch Normalization}
\label{sec:sbn}
Similar to Batch Normalization (BN) \cite{Ioffe2015:BN}, Sequence-wise Batch Normalization (SBN) \cite{SBN2015} normalizes the input or the output of a recurrent neural network (RNN) across the batch and the temporal dimension. Let $\Phi \in \mathcal{R}^{B\times T \times P}$ be an order-3 tensor where $B$ is the batch size, $T$ the time steps, and $P$ the dimension of the input vector at each time step. We write $\Phi_{b, t, p}$ for the scalar elements in the tensor. Assuming the number of unpadded elements in the tensor is $M$, we compute the mean $\mu_p$, the variance $\sigma_p^2$, and the normalized tensor $\tilde{\Phi}$ as
\begin{align}
    \mu_p & = \frac{1}{M} \sum_{b=1}^{B} \sum_{t=1}^{T} \Phi_{b, t, p} \\
\sigma_p^2 & = \frac{1}{M} \sum_{b=1}^{B} \sum_{t=1}^{T} (\Phi_{b, t, p} - \mu_p)^2 \\
    \text{SBN}(\Phi) & = \tilde{\Phi}_{b, t, p} = \gamma_p \frac{\Phi_{b, t, p}-\mu_p}{\sqrt{\sigma_p + \epsilon}} + \beta_p
\end{align}
where $\epsilon$ is a small constant to prevent numeric instability. $\gamma_p$ and $\beta_p$ are per-dimension scaling and bias factors that restore network expressiveness after normalization. 

To align the distributions of the video and the textual modalities, we apply sequence-wise batch normalization at the output of the respective LSTMs. Though the effects of normalization on convolutional and feedforward networks have been extensively studied (e.g.,  \cite{hoffer2018norm,Bjorck:BN2018,Santurkar2018,arora2018theoretical,cai2018quantitative}), the recurrent variants (e.g., \cite{BN-RNN:2016,ren2016normalizing,RMSNorm}) are less well understood.

\subsection{Layer Normalization}

Layer Normalization (LN) \cite{ba2016:LayerNorm} is another common normalization technique for RNNs. Instead of normalizing across batch and time steps, it normalizes across the dimensions of the input vectors. 

\begin{equation}
    \mu_{b,t} = \frac{1}{P} \sum_{p=1}^{P} \Phi_{b, t, p}
\end{equation}  

\begin{equation}
    \sigma_{b,t}^2 = \frac{1}{P} \sum_{p=1}^{P} (\Phi_{b, t, p} - \mu_{b,t})^2 \\
\end{equation}

\begin{equation}
    \text{LN}(\Phi)  = \tilde{\Phi}_{b, t, p} = \gamma_p \frac{\Phi_{b, t, p}-\mu_{b,t}}{\sqrt{\sigma_{b,t} + \epsilon}} + \beta_p
\end{equation}
Preliminary experiments show that directly applying layer normalization to the LSTM outputs performed poorly. Instead, we employ the popular configuration from PyTorch FastRNN\footnote{\url{https://github.com/pytorch/pytorch/blob/master/benchmarks/fastrnns/custom_lstms.py}}, which applies LN to the internal gates and the cell rather than the output. 
\begin{equation}
    \begin{bmatrix}
\bm i_t^\top, \bm f_t^\top, \tilde{\bm c}_t^\top, \bm o_t^\top 
\end{bmatrix}^\top = \text{LN}\left(
W_x \bm x_t \right) + \text{LN}\left(
W_h \bm h_{t-1} \right)
\end{equation}
\begin{equation}
\bm c_t = \text{LN}\left(\sigma(\bm f_t) \otimes \bm c_{t-1} + \sigma(\bm i_t) \otimes \tanh(\tilde{\bm c}_t\right)) 
\end{equation}
The hidden state computation (Eq. \ref{eq:lstm-h}) remains unchanged. In the experiments, we find that LN provides good performance elevation, but not as effectively as SBN. 

\subsection{Random Projection}
\label{sec:rand_proj}
To reduce input dimensionality and trainable parameters, we adopt a simple random projection technique by Achlioptas \cite{Achlioptas2001:RP}. Any input feature $\bm x \in \mathcal{R}^D$ is projected to $P$ dimensions by multiplying with a random matrix $R \in \mathcal{R}^{P\times D}$, whose elements are random sampled as
\begin{equation}
    R_{ij} = \begin{cases}
    \sqrt{3} , & \text{with probability } 1/6 \\
    0,  & \text{with probability } 2/3 \\
    -\sqrt{3},  & \text{with probability } 1/6 \\
\end{cases}
\end{equation}
This projection has been shown to preserves distances between input vectors up to a constant factor $\sqrt{P}$ \cite{Achlioptas2001:RP}.

\section{Experiments}
In this section, we present the experimental results and analyze the effects of the three techniques of LARS, SBN, and random projection.\footnote{Code is available at \url{https://github.com/RubbyJ/Data-efficient-Alignment}.}
 
\subsection{Dataset and Performance Metrics}
The  YouTube Movie Summary  dataset contains 93 short videos, each lasting about a few minutes for a total of 6.7 hours. The videos contain narration of the movie stories and selected clips from the original movie. The dataset was annotated at the sub-sentence level; a sentence can be broken down into multiple snippets and aligned with different video segments. We follow the original split and use 66 videos for training, 12 for validation, and 15 for testing. 

In preprocessing, we follow \cite{NeuMatch2018} and use a threshold-based method\footnote{\url{https://github.com/Breakthrough/PySceneDetect}} for scene boundary detection to segment a video into a sequence of video clips. We set the threshold hyperparameter to 20 and the minimum length of a clip to 5 frames. We use the ground-truth chunking of sentences. Due to the existence of long sequences, we split training sequences so that each training sample contains 100 actions or less in order to improve training efficiency. The validation and testing sequences are unaffected.  

Performance on this task is measured using video alignment accuracy and text alignment's overlap with ground truth. Since one video clip can match with at most one text snippet, video alignment accuracy is computed as the total length of correctly aligned video clips over the length of all clips. A text snippet can match with multiple video clips, so we compute the temporal intersection over union (IoU) of the predicted and the ground-truth clips matched to one snippet. Due to less-than-optimal video segmentation, the maximum training accuracies for video and text are 98.2\% and 93.6\% respectively.

\subsection{Experimental Setup}

We extract video and text features in the following manner. Following \cite{Anderson2017}, we extract features from the central frame of each video clip using a Faster-RCNN \cite{fasterrcnn2015} trained on the Visual Genome dataset \cite{visualgenome2016}. For a text snippet, we extract 768-dimensional sentence embedding from the BERT-Base model \cite{devlin2018bert}. 
After that, both features are reduced to 300 dimensions using either random projection (Section \ref{sec:rand_proj}) or trainable dense layers. We normalized features to zero mean and unit variance before the trainable portion of the network. 

Like the original NeuMATCH \cite{NeuMatch2018}, we use 2 layers of LSTM in all stacks with 300 hidden dimensions for Video/Text Stacks, 20 for Matched Stack, and 8 for Action Stack. Additionally, we append ten positional features, such as the number of elements remaining in Video/Text Stacks, elements already matched, their ratios, and their reciprocals. 

All experiments train for 350 epochs and use a batch size of 32. 
We use Adam as the optimizer and LARS is added to Adam for some networks. The norm of gradients is capped at 2. 
The learning rate is halved whenever training loss does not improve for 10 epochs. The best performing model, RP+LARS+SBN, adopts an initial global learning rate of 0.007 and label smoothing \cite{Szegedy_2016} of 3\%. For comparison against pretraining on other multimodal data, we also use unimodal feature extractors from LXMERT \cite{tan2019lxmert} to generate features for the central frames of the video clips and the text snippets.  See the supplemental material for more experimental details.

\begin{figure}
    \centering
    \includegraphics[scale=0.45]{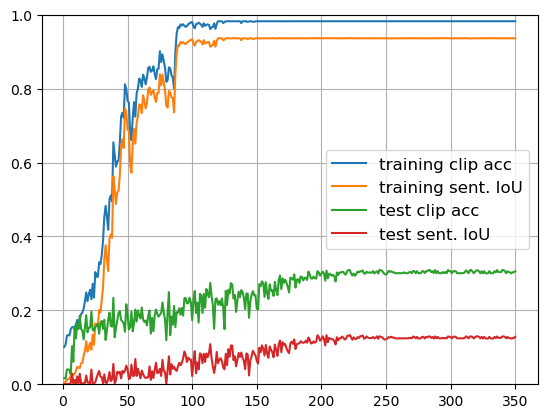}
    \caption{The training trajectory of the RP+LARS+SBN network. The x-axis indicates the number of epochs and the y-axis indicates accuracy / IoU. }
    \label{fig:training}
\end{figure}

\subsection{Main Results}
In Table \ref{table:main-results}, we present the performance of different networks with and without pretraining on the LSMDC dataset. Naively omitting pretraining leads to severe overfitting; the validation set performance peaks at epoch 56, after which the performance drops. At epoch 350, there is a 3.6\% performance gap in text IoU in comparison to \cite{NeuMatch2018}. The improvement in video accuracy is likely due to better input features. When we choose epoch 56 using validation performance (\emph{i.e.} early stopping), the test performance gap can be reduced to 2.4\%. 
When RP, LARS, and SBN are applied, we achieve the best performance of 30.6\% video accuracy and 12.8\% text IoU. The three techniques achieve strong regularization effects so that testing performance remains stable toward the end of the training (see Figure \ref{fig:training}). With strong LXMERT features, we can obviate the need for pretraining on LSMDC, but this baseline is still inferior to the network using RP+LARS+SBN, suggesting RP+LARS+SBN may curtail overfitting in the sequence encoders. 
In the next few subsections, we present a detailed ablation study on the effects of the three techniques. 

\begin{figure*}[t]
\begin{subfigure}{.5\textwidth}
    \centering
    \includegraphics[scale=0.46]{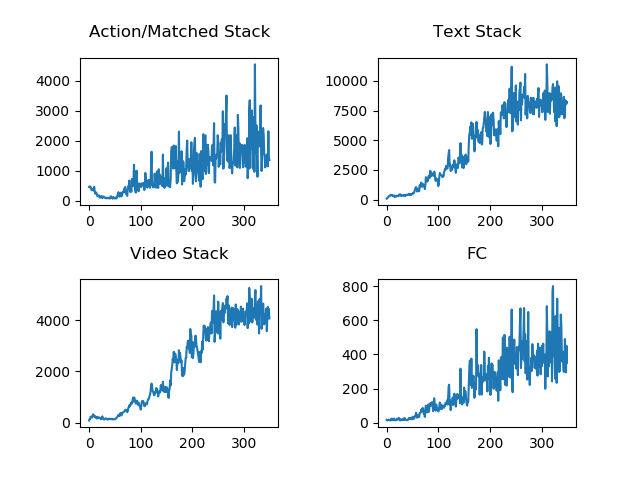}
    \caption{Adam + RP + SBN}
\end{subfigure}
\begin{subfigure}{.5\textwidth}
    \centering
    \includegraphics[scale=0.46]{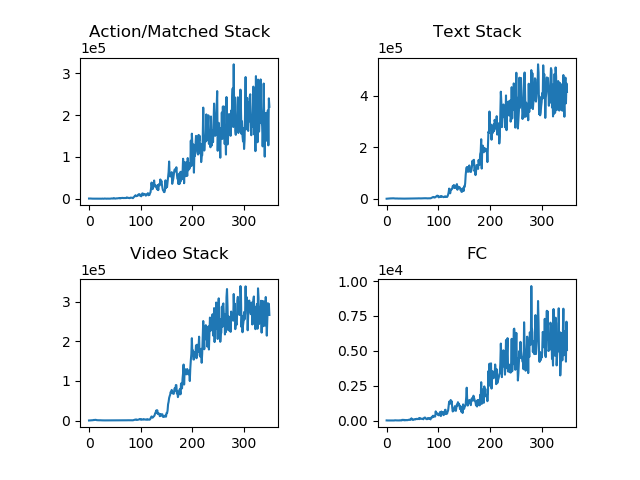}
    \caption{LARS + RP + SBN}
\end{subfigure}
\caption{The ratio $\frac{\|\bm w\|_2}{\|\nabla_{w}\mathcal {L}\|_2}$ between the parameter norm and the gradient norm in four network components. A greater ratio indicates the model is closer to a local minimum.}
\label{fig:gradien-ratio}
\end{figure*}

\begin{table}[t]
\centering
\begin{tabular}
{@{}lcc@{}}
 \toprule
 \textbf{Model} & \textbf{Video Acc.} & \textbf{Text IoU} \\ 
 \midrule
 NeuMATCH w/ pretraining \cite{NeuMatch2018} & 12.0 & 10.4 \\ 
 Our pretraining & 26.8 & 10.3 \\
 No pretraining & 24.2 & 6.8 \\
 No pretraining + early stopping & 24.9 & 8.0 \\
 No pretraining + LXMERT features & 28.5 & 10.8 \\
 No pretraining + RP+LARS+SBN & \textbf{30.6} & \textbf{12.8} \\ 
 \bottomrule
\end{tabular}
 \vspace{0.05in}
 \caption{Test performance when NeuMATCH is trained with or without pretraining on LSMDC and the regularization techniques RP, LARS, and SBN. }
\label{table:main-results}
\end{table}

\subsection{Effects of LARS}
\label{sec:lars-analysis}
In order to study the regularization effects of the techniques, in this and all subsequent experiments, we train all models to maximum training accuracy under the current video segmentation (98.2\% and 93.6\% for video and text respectively). As a result, the test accuracy directly reflects the generalization gap at convergence. As the dataset is small, we find results from early stopping to have high variance and unreliable.

\begin{table}[t]
\centering
\begin{tabular}
{@{}lcc@{}}
 \toprule
  \textbf{Optimizer} & \textbf{Video Acc.} & \textbf{Text IoU} 
 \\
 \midrule 
  Adam & 21.7 & 8.0 \\
  Adam + warm-up & 24.7 & 9.3 \\
  LARS  & \textbf{30.6} & \textbf{12.8} \\ 
 \bottomrule
\end{tabular}
 \vspace{0.05in}
 \caption{Comparing LARS to Adam and learning rate warm-up on test performance. All model use randomly projected features and SBN.}
\label{table:optimizer}
\end{table}

\begin{table*}
\centering
\begin{tabular}
{lccccc}
 \toprule
 \textbf{Model} & \textbf{Video Stack} & \textbf{Text Stack} & \textbf{$\frac{\text{Text Stack}}{\text{Video Stack}}$} & \textbf{Action \& Matched Stacks} & \textbf{FC} \\ 
 \midrule
 RP + Adam + SBN & 4263.5 & 8075.0 & 1.9 & 1711.9 & 432.1 \\
 RP + Adam + warm-up + SBN & 10378.1 & 23305.1 & 2.2 & 8228.3 & 2120.8 \\
 RP + LARS + SBN & $2.7\times 10^{5}$ & $4.1\times 10^{5}$ & 1.5 & $1.9\times 10^{5}$ & 5604.5 \\
 RP + Adam + 2$\times$LN & $2.6\times10^{6}$ &$5.5\times10^{6}$ & 2.1 & $1.1\times10^{6}$ & $7.4\times10^{4}$ \\
 RP + LARS + 2$\times$LN & $3.5\times10^{6}$ &$6.4\times10^{6}$ & 1.8 & $1.4\times10^{6}$ & $9.0\times10^{4}$ \\
 RP + Adam + 4$\times$LN & $2.3\times10^{6}$ & $3.8\times10^{6}$ & 1.7 & $7.3\times10^{5}$ & $7.8\times10^{4}$ \\
 RP + LARS + 4$\times$LN & $1.3\times10^{7}$ & $2.1\times10^{7}$ & 1.6 & $1.0\times10^{7}$ & $2.5\times10^{5}$ \\
 \bottomrule
\end{tabular}
 \vspace{0.05in}
 \caption{Means of the gradient norm ratio $\frac{\|\bm w\|_2}{\|\nabla_{w}\mathcal
{L}\|_2} $ in the last 20 training epochs for different models and model components.}
 \label{table:sgd_ratio}
\end{table*}

\begin{figure*}[t]
\centering
\includegraphics[scale=0.76]{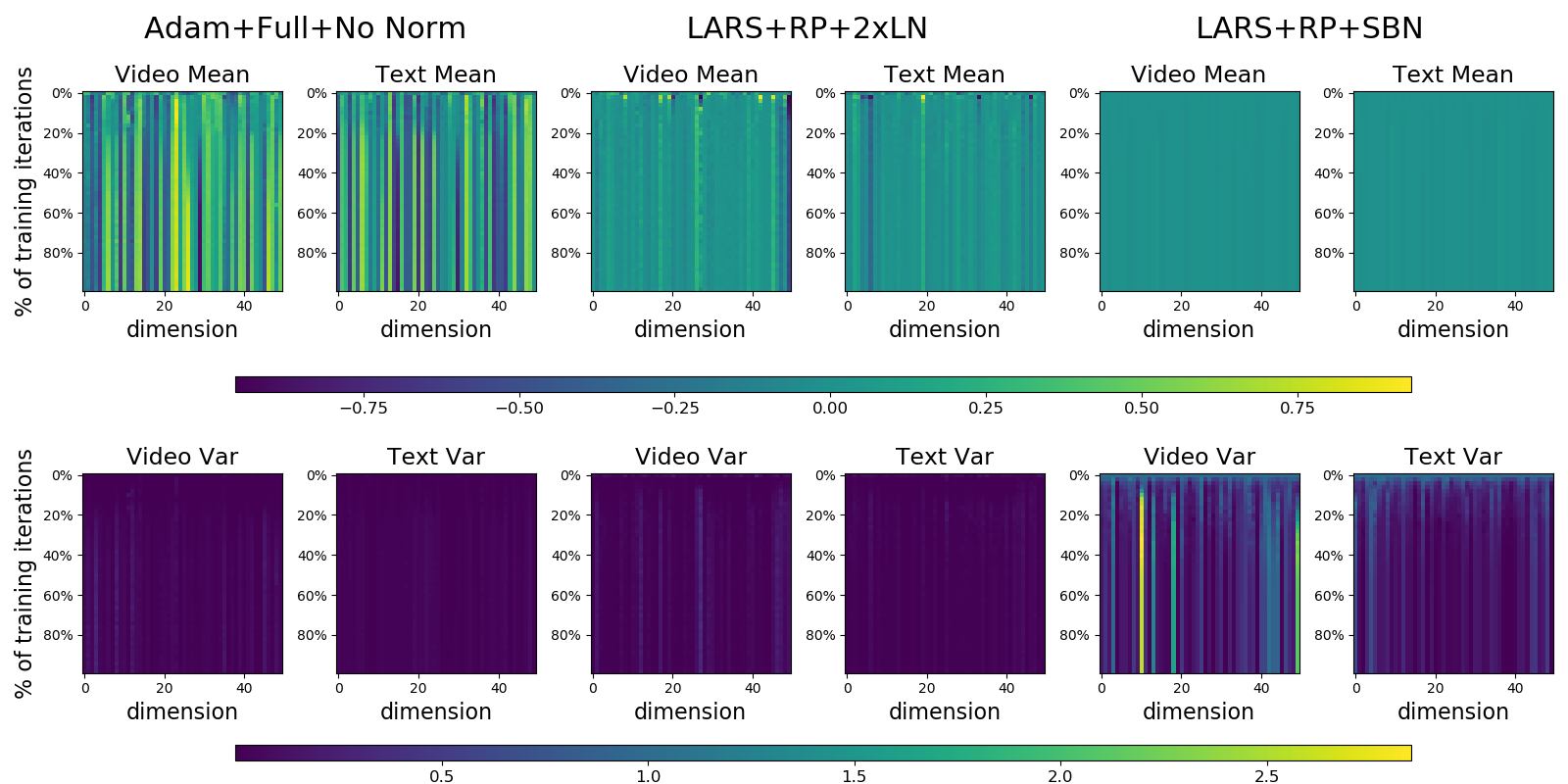} 
\caption{The mean and variance distributions of the internal video and text feature, which are the outputs of the Video and Text Stacks. We plot only the first 50 dimensions throughout the training process and compare SBN against the network without any normalization.}
\label{fig:mean_var}
\end{figure*}

Table \ref{table:optimizer} shows the effects of LARS on test performance. Using the Adam optimizer alone leads to severe overfitting (21.7\% video accuracy and 8.0\% text IoU). %
LARS reduces the generalization gap by 8.9\% for video accuracy and 4.8\% for text IoU. 

We further inspect the ratio between the norm of model parameters $\|\bm w\|_2$ and the gradient $\|\nabla_{\bm w}\mathcal{L}\|_2$ in the Video Stack, the Text Stack, the Action/Matched Stacks, and the two FC layers before the final softmax. A higher ratio indicates the network is closer to a local minimum. Results are shown in Figure \ref{fig:gradien-ratio} and Table \ref{table:sgd_ratio}.

Comparing the three models in Table \ref{table:optimizer}, we observe that LARS provides substantial benefits in the optimization of the network; LARS always achieves higher ratios and hence better convergence than Adam + warm-up and Adam. In particular, with both RP and SBN, applying LARS causes a 20-fold increase of the ratio over Adam+warm-up for the Action / Matched Stack. This indicates these two stacks converge significantly better with LARS than without. 
We further note that the gradient norm ratios under LARS increase sharply after 200 epochs.  
Observing the training trajectory of RP+LARS+SBN in Figure \ref{fig:training}, the network is getting very close to 100\% training accuracy at this time, yet further optimization is still beneficial. This indicates that LARS provides good optimization even when the network approaches the minimum and the gradient becomes small. 

It may seem counter-intuitive that better optimization can narrow the generalization gap. We contend that LARS contributes to generalization because it aligns the magnitudes of the gradient updates and allows the four stacks to train at similar speed. As a result, the FC layers do not overfit to any particular stack. This argument is supported by two facts. First, the Action/Matched Stacks are much better optimized under LARS. Second, compared to Adam, LARS always reduces the relative gap between the $\|\bm w\|_2 / \|\nabla_{\bm w}\mathcal{L}\|_2$ ratios of the Video Stack and the Text Stack. For example, with RP+Adam+SBN, the ratio of the Text Stack is 1.9 times the ratio of the Video Stack. With RP+LARS+SBN, this reduces to 1.5. This suggests the optimization of the two modalities is better aligned with LARS than without.

\subsection{Effects of Normalization}
\label{sec:sbn-analysis}
We compare SBN against the popular layer-normalized LSTMs. In the 2$\times$LN version, we apply LN to the Video and Text Stacks. In the 4$\times$LN version, all four stacks are layer-normalized.  

Table \ref{table:norm} shows that when LARS is applied, SBN outperforms 4$\times$LN, which slightly outperforms 2$\times$LN. Observing Table \ref{table:sgd_ratio}, we notice that LARS+LN can push the gradient norm ratios even higher than LARS+SBN.  
SBN creates noise in the gradient as data points in the same batch are randomly sampled. The noise may have prevented extremely close convergence or falling into sharp minima \cite{smith2018bayesian}, thereby providing some regularization effects. 

With Figure \ref{fig:mean_var}, we additionally examine the mean and variance of the features provided by the Video and Text Stacks when we apply SBN, 2$\times$LN, and no normalization. Under SBN, the means are very close to zero but the variances are much higher than the other versions. This suggests the features are more informative under SBN because they change more for different input. Even though the SBN contains a bias factor that may shift the mean away from zero, the network does not learn to do so.
\begin{table}[t]
\centering
\begin{tabular}{p{5.5em} c c c}
 \toprule
 \textbf{Normalization} & \textbf{Optimizer} & \textbf{Video Acc.} & \textbf{Text IoU}\\ 
 \midrule
 SBN & Adam & 21.7 & 8.0 \\
 2$\times$LN & Adam & 23.3 & 8.2 \\
 4$\times$LN & Adam & 23.6 & 7.5 \\
 SBN & LARS &  \textbf{30.6} & \textbf{12.8} \\ 
 2$\times$LN & LARS & 25.3 & 8.8 \\
 4$\times$LN & LARS & 25.9 & 8.7 \\
 \bottomrule
\end{tabular}
 \vspace{0.05in}
 \caption{Comparison of different normalization techniques. Random projection is used for all models in this table.}
\label{table:norm}
\end{table}

\begin{table}[t]
\centering
\begin{tabular}
{@{}lccc@{}}
 \toprule
 \textbf{Feature / Norm} 
 & \textbf{Video Acc.} & \textbf{Text IoU} & \textbf{\# Parameters}\\ 
 \midrule
Full + SBN  & 27.6 & 9.9 & 4.25M \\ 
 RP + SBN  & \textbf{30.6} & \textbf{12.8} & 3.41M \\ 
 Full + 2$\times$LN  & 25.4 & 8.3 & 4.26M \\
 RP + 2$\times$LN  &  25.3 & 8.8 & 3.42M \\
 Full + 4$\times$LN  & 26.9 & 9.9 & 4.26M \\
 RP + 4$\times$LN  & 25.9 & 8.7 & 3.42M \\
 \bottomrule
\end{tabular}
 \caption{Comparison between full and RP features and the number of trainable parameters. All models use LARS. }
\label{table:feature}
\end{table}

\subsection{Effects of Random Projection}
\label{sec:rp-analysis}
Table \ref{table:feature} compares models using  features in the original dimensions (Full) and randomly projected (RP) features. To create fair comparisons in this experiment, the only difference within each Full-versus-RP pair is whether the FC layers before the Stacks are trainable or fixed random projection matrices. The models use different normalization techniques, SBN, 2$\times$LN, and 4$\times$LN and are always optimized with LARS. RP features usually obtain better performance over the model using full features, though full features perform better for 4$\times$LN. The regularization efforts of RP can be attributed to the fact that RP reduces the number of trainable parameters by 19\% in the models we tested.

\section{Conclusions}

Establishing fine-grained correspondences  between the vast reservoir of parallel movie videos and screenplays can empower many practical applications. A major roadblock is the limited availability of annotated data. The large LSMDC dataset deviates from realistic situations; the YMS dataset is more realistic but 20 times smaller. 
With this paper, we aim to improve data efficiency of sequence alignment networks, which helps in removing the need to pretrain on the LSMDC dataset. 

The three techniques, namely layer-wise adaptive rate scaling, sequence-wise batch normalization, and random projection of input features, allow the network to beat the previous best result, which was pretrained on multimodal data. Experiments suggest that LARS helps in aligning the pace of learning in different network components. We believe this work represents a step toward data-efficient multimodal learning. 

\noindent\small\textbf{Acknowledgement} This work was supported in part by NSFC Project (62076067), and Science and Technology Commission of Shanghai Municipality Project (\#19511120700). Boyang Li is partially supported by Alibaba Group through Alibaba Innovative Research (AIR) Program and Alibaba-NTU Singapore Joint Research Institute (Alibaba-NTU-AIR2019B1), Nanyang Technological University, Singapore.

{\small
\bibliographystyle{ieee_fullname}
\bibliography{egbib}
}

\clearpage
\appendix
  \renewcommand{\appendixname}{Appendix~\Alph{section}}
\section{Details on Experiment Setup}
In this supplementary material, we give the experiment details in Section 4 in our main paper. 

All the experiments share the same batch size 32 and the same learning rate scheduler, which halves the learning rate if training loss does not improve in the last ten epochs. The initial global learning rates for different experimental setups have been tuned for performance. 

\subsection{The ``Our Pretraining'' Baseline}
This baseline employs Faster RCNN and BERT features, which were not used by \cite{NeuMatch2018}. With this baseline, we first pretrain NeuMATCH on the large LSMDC dataset and finetune it on YMS. For pretraining, the initial learning rate is set to $1\times10^{-3}$ and weight decay to $1\times10^{-6}$ and dropout to 0.1 for all fully-connected layers. During pretraining, we apply early stopping using the validation set, which stopped the training at Epoch 47. Finetuning on YMS uses the initial learning rate of $5\times10^{-4}$ and dropout of 0.1 for all fully-connected layers. Weight decay is not used during finetuning. We use the Adam optimizer for both pretraining and finetuning. 

\subsection{The ``LXMERT Features'' Baseline}

As another baseline, we extract features using the unimodal encoders from LXMERT. The full LXMERT model first feeds inputs through respective unimodal encoders, followed by a cross-modal encoder. As we do not have image-text correspondence before encoding, we simply use the pretrained unimodal encoders. We perform mean pooling over the RoI features extracted by the image encoder and over the word-level features from the text encoder. 

In training this baseline, the initial learning rate is $3\times10^{-4}$ with weight decay of $1\times10^{-7}$ and dropout of 0.3 for all fully-connected layers. The Adam optimizer is used. 

\subsection{Ablation Experiments}
\begin{table}[h]
    \centering
    \begin{tabular}{@{}lccc @{}}
        \toprule
        \textbf{Feature} & \textbf{Optimizer} & \textbf{Normalization} & \textbf{Initial learning rate}  \\
         \midrule
        Full & Adam & — & $1\times10^{-3}$\\
        Full & LARS & SBN & $5\times10^{-3}$\\
        Full & LARS & 2$\times$LN & $4\times10^{-3}$\\
        Full & LARS & 4$\times$LN & $5\times10^{-3}$\\
        RP & Adam & SBN & $5\times10^{-4}$ \\
        RP & Adam & 2$\times$LN & $1\times10^{-2}$ \\
        RP & Adam & 4$\times$LN & $7\times10^{-3}$ \\
        RP & LARS & SBN & $7\times10^{-3}$ \\
        RP & LARS & 2$\times$LN & $5\times10^{-3}$ \\
        RP & LARS & 4$\times$LN & $8\times10^{-3}$ \\
         \bottomrule
    \end{tabular}
    \caption{Initial global learning rates for ablation experiments. All models use a label smoothing regularization of 3\%.}
    \vspace{0.05in}
    \label{table:lrs}
\end{table}

Table \ref{table:lrs} gives the experiment details from Section 4.4-4.6. All these experiments apply a label smoothing regularization to the distribution of ground truth label. The label smoothing hyperparameter $\epsilon=0.03$. Weight decay and dropout are not used.
In the setup of RP+Adam warm-up+SBN, we linearly increase the learning rate for 5 epochs from $5 \times 10^{-5}$ to $5 \times 10^{-4}$ without any label smoothing.

For completeness, we formally define label smoothing. With the hyperparameter $\epsilon$, label smoothing modifies the ground-truth class probability to $(1-\epsilon)$ and evenly distributes $\epsilon$ among the rest of the classes. We use the modified probability vector as the target in cross-entropy loss. 

More formally, we can denote the ground-truth labels as 
$[y_1, \ldots, y_k, \ldots, y_K]$, $y_k \in \{0, 1\}$. When the true class is $k$, we set $y_k = 1$ and rest of the labels as $0$. With label smoothing, the new labels are set to
$$ y_k^{smooth} = (1-\epsilon)y_k + \frac{\epsilon}{K} $$
We choose $\epsilon=0.03$ for all the experiments in Table \ref{table:lrs}.  
\\

\end{document}